\documentclass[10pt,twocolumn,letterpaper]{article}

\usepackage{iccv}
\usepackage{times}
\usepackage{epsfig}
\usepackage{graphicx}
\usepackage{amsmath}
\usepackage{amssymb}
\usepackage[linesnumbered,algoruled,boxed,lined]{algorithm2e}
\usepackage{adjustbox}
\usepackage{comment}
\usepackage{multirow}
\usepackage{import}
\usepackage[noend]{algpseudocode}
\usepackage[dvipsnames]{xcolor}
\usepackage{soul}

\usepackage[pagebackref=true,breaklinks=true,letterpaper=true,colorlinks,bookmarks=false]{hyperref}

\iccvfinalcopy 

\def\iccvPaperID{31} 

\ificcvfinal\fi
\begin{document}

\title{Attention routing between capsules}

\author{Jaewoong Choi \hspace{0.5cm} Hyun Seo \hspace{0.5cm} Suii Im \hspace{0.5cm} Myungjoo Kang\\
Seoul National University\\
{\tt\small \{chjw1475, hseo0618, a5828167, mkang\}@snu.ac.kr}
}

\maketitle
\ificcvfinal\thispagestyle{empty}\fi

\begin{abstract}
In this paper, we propose a new capsule network architecture called Attention Routing CapsuleNet (AR CapsNet). We replace the dynamic routing and squash activation function of the capsule network with dynamic routing (CapsuleNet) with the attention routing and capsule activation. The attention routing is a routing between capsules through an attention module. The attention routing is a fast forward-pass while keeping spatial information. On the other hand, the intuitive interpretation of the dynamic routing is finding a centroid of the prediction capsules. Thus, the squash activation function and its variant focus on preserving a vector orientation. However, the capsule activation focuses on performing a capsule-scale activation function.

We evaluate our proposed model on the MNIST, affNIST, and CIFAR-10 classification tasks. The proposed model achieves higher accuracy with fewer parameters ($\times$0.65 in the MNIST, $\times$0.82 in the CIFAR-10) and less training time than CapsuleNet ($\times$0.19 in the MNIST, $\times$0.35 in the CIFAR-10). These results validate that designing a capsule-scale operation is a key factor to implement the capsule concept.

Also, our experiment shows that our proposed model is transformation equivariant as CapsuleNet. As we perturb each element of the output capsule, the decoder attached to the output capsules shows global variations. Further experiments show that the difference in the capsule features caused by applying affine transformations on an input image is significantly aligned in one direction.

\end{abstract}

\begin{figure*}[t]
\begin{center}
\includegraphics[width=0.95\linewidth]{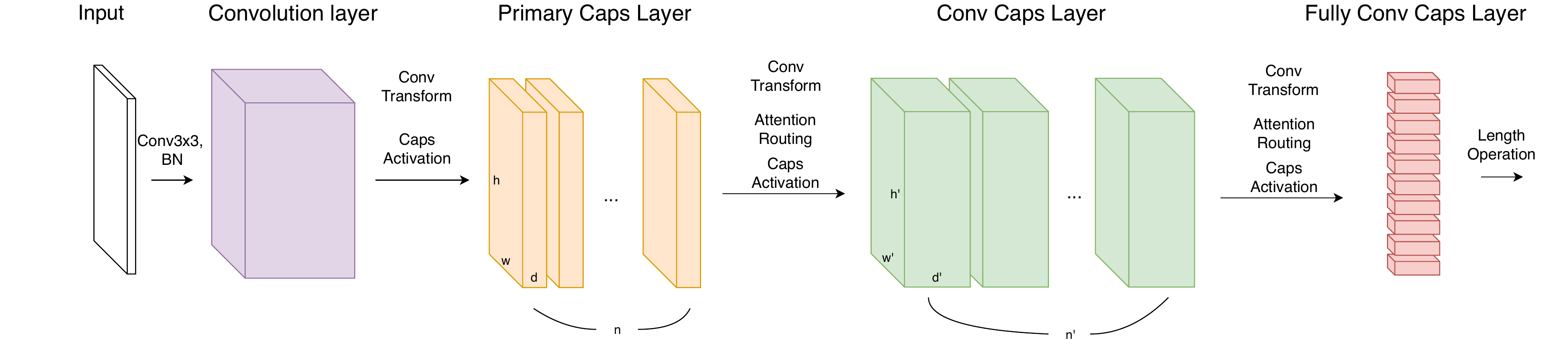}
\end{center}

   \caption{Overview of AR CapsNet. AR CapsNet is composed of primary caps layer, conv caps layer, and fully conv caps layer. BN denotes the batch normalization. Conv Transform and Caps Activation denotes the convolutional transform and capsule activation respectively.}
\label{fig:model}
\end{figure*}

\section{Introduction}
Convolutional neural networks(CNNs) have had much success in computer vision tasks \cite{AlexNet} \cite{VGGNet} \cite{ResNet} \cite{DenseNet}. The convolutional layer is an effective method to extract local features due to its local connectivity and parameter sharing with spatial location. However, the convolutional layer has a limited ability to encode a transformation. For example, if the convolutional layer is combined with a max-pooling layer, the extracted feature is local translation invariant. As CNN models become deeper \cite{ResNet} \cite{HighwayNetwork}, the receptive field of each feature is getting larger. Then, the information loss from the translation invariance also increases. 

To overcome the transformation invariance of CNNs, the transforming autoencoder \cite{transformingautoencoder} uses the concept of "capsule". A capsule is a vector representation of a feature. Each capsule not only represents a specific type of entity but also describes how the entity is instantiated, such as precise pose and deformation. In other words, the capsules are transformation equivariant.

The CapsuleNet \cite{capsulenet} is a novel method that implements the idea of the capsules. By introducing the dynamic routing algorithm and squash activation function \ref{eq:squash}, CapsuleNet uses vector-output capsules as a basic unit instead of scalar-output features.
\begin{equation}\label{eq:squash}
    \textrm{squash}(\mathbf{s_j}) = \frac{||\mathbf{s}_{j}||^2}{1+ ||\mathbf{s}_{j}||^2}
                                        \frac{\mathbf{s}_{j}}{||\mathbf{s}_{j}||}
\end{equation}
where $\mathbf{s_j}$ is a pre-activation capsule. However, CapsuleNet has a room for development. The number of parameters of CapsuleNet is much larger than that of comparable performance CNN-based models. Also, the dynamic routing is an iterative process. The reported accuracy of CapsuleNet on the benchmark datasets like CIFAR-10 is inferior to state-of-the-art CNN models. \cite{complex}

In this paper, we propose a convolutional capsule network architecture comprised of building blocks of CNNs. We substitute the dynamic routing and squash capsule-activation function of CapsuleNet\cite{capsulenet} with attention routing and capsule activation. In the attention routing, the log probabilities of agreement coefficients between the $l$th layer and the $(l+1)$th layer are learned by a scalar-product between the capsules of the $l$th layer and the kernel of convolution. The kernel of convolution serves as an approximation of the reference vector to perform routing. By replacing an iterative process of the dynamic routing with forward-pass convolution, the attention routing is fast while maintaining spatial information. 
Two important properties of squash activation function \ref{eq:squash} is that the squash activation function preserves a vector orientation and is a capsule-wise activation function, not an element-wise activation function such as $\textrm{ReLU}$ or $\tanh$. 
The dynamic routing is an unsupervised algorithm to find a centroid-like output capsule of the prediction capsules. Therefore, the squash activation function and its variant \ref{eq:squash_variant} \cite{complex} focus on preserving a capsules orientation.
\begin{equation}\label{eq:squash_variant}
    \textrm{squash variant}(\mathbf{s_j}) = \left( 1-  \frac{1}{\exp{( ||\mathbf{s}_{j}||}) } \right)
                                        \frac{\mathbf{s}_{j}}{||\mathbf{s}_{j}||}
\end{equation}
However, we focus on the capsule-wise operation rather than preserving orientation. The capsule activation performs an affine transform on the capsules and then applies an element-wise activation function. The capsules on the same capsule channel share parameters used in the affine transformation. Thus, the capsules on the same capsule channel are mapped to the same feature space, and the operation is parameter efficient. Therefore, the capsule activation is a capsule-wise function that does not preserve a vector orientation. Since the capsule activation applies a nonlinear transformation to a linear combination of the prediction capsules, parametrizing the routing process through the attention routing is compatible. We refer to our proposed model as \textit{Attention Routing CapsuleNet (AR CapsNet).}

We evaluate the AR CapsNet on three datasets (MNIST, affNIST, and CIFAR-10). The AR CapsNet significantly outperforms CapsuleNet in the affNIST and CIFAR-10 classification task and shows a comparable performance in the MNIST dataset while being faster and using less than half parameters than CapsuleNet. Moreover, the AR CapsNet preserves the transformation equivariant property of CapsuleNet. As we perturb each element of the output capsule, the decoder attached to the output capsules shows global variations as in \cite{capsulenet}. Further experiment showed that the affine transformations on an input image cause the feature capsules to change in the significantly aligned direction. From these experiments, we prove that the AR CapsNet encodes an affine transformation on the input image in some basis of capsule space. In addition, our proposed architecture is constructed in a convolutional manner so that it can be easily extended to a deeper network structure. 

\subsection{Contribution}
\begin{itemize}
    \item We propose a new architecture called AR CapsNet by introducing two modifications to the CapsuleNet \cite{capsulenet}. These modifications are the \textit{attention routing} and \textit{capsule activation.}
    
    \item The capsule activation expands the concept of the existing capsule-wise activation functions such as the squash activation. The capsule activation performs an orientation-nonpreserving transform on the pre-activation capsules. The performance of the AR CapsNet demonstrates that the transformation equivariant features can be extracted even if the routing process is not restricted to the clustering approach and the capsule activation is not limited to the normalization.
    
    \item The AR CapsNet shows better results on the affNIST, and CIFAR-10 classification tasks and comparable results on the MNIST classification task while using much smaller parameters than CapsuleNet. Also, the AR CapsNet preserves the transformation equivariant property of the CapsuleNet. As we perturb each element of the output capsule, the decoder attached to the output capsule shows global variation as in \cite{capsulenet}.
    
    \item To investigate the transformation equivariance further, we suggest a new experiment. We observe the difference in the output capsule caused by applying transformations on an input image. In the AR CapsNet, these difference vectors are significantly aligned compared to a set of random vectors. These results demonstrate that transformation on an input image is encoded in some basis vector.
    
\end{itemize}

\section{Related Works}
The CNN models that consist of convolutional layers and max-pooling layers have a local translation invariance. To overcome transformation invariance, CapsuleNet \cite{capsulenet} uses vector-output capsules and the dynamic routing in place of scalar-output features and max-pooling. By demonstrating that the dimension perturbation of digit capsules leads to a global transformation of the reconstruction image, CapsuleNet claims to have transformation equivariance. 

A number of methods to improve the performance of CapsuleNet have been proposed in \cite{EMrouting} \cite{wang2018optimization} \cite{SpectralCapsulenet} \cite{SegCaps} \cite{NeuralNetworkEncapsule}. In \cite{wang2018optimization}, they interpreted the routing-by-agreement process as an optimization problem of minimizing clustering loss. They proposed another routing process from the point of view of clustering. Their approach achieved better results on an unsupervised perceptual grouping task compared to \cite{capsulenet}. The matrix capsules with EM routing \cite{EMrouting} proposed another routing method called EM routing. The EM routing measures compatibility between matrix capsules by clustering matrix capsules through Gaussian distributions. The matrix capsules with EM routing achieved the state-of-the-art performance on a shape recognition task using the smallNORM dataset. The spectral capsule networks \cite{SpectralCapsulenet} is a variation of \cite{EMrouting}. Spectral capsule networks use a singular value to compute the activation of each capsule instead of the logistic function in \cite{EMrouting}. Spectral capsule networks achieved better performance on a diagnosis dataset compared to \cite{EMrouting} and deep GRU networks while showing faster convergence compared to \cite{EMrouting}.

The SegCaps \cite{SegCaps} applied a capsule network to the object segmentation task. The SegCaps introduced two modifications to the CapsuleNet and devised the concept of deconvolutional capsules from these modifications. The two modifications are the locally connected dynamic routing and the sharing of transformation matrices within the same capsules channel. The sharing of transformation matrices is equivalent to the convolutional transform of our conv caps layer except for the addition of biases. 
The EncapNet \cite{NeuralNetworkEncapsule} performs a one-time pass approximation of the routing process by introducing two branches. The master branch extracts a feature from the locally connected capsules as in \cite{SegCaps} and the aide branch combines information from all the remaining capsules. Also, they introduced a Sinkhorn divergence loss which works as a regularizer. The EncapNet achieved competitive results on CIFAR-10/100, SVHN, and a subset of ImageNet. 

Our proposed model uses attention architecture as a routing algorithm. The attention architecture learns a compatibility function between low-level features and high-level features. In \cite{BahdanauAttention}, the output of attention architecture is a weighted sum of input features, and the weights are the compatibilities based on the input features and the RNN hidden state. The compatibility function is a feedforward neural network with a softmax activation function. In \cite{LuongAttention}, they experimented on various kinds of attention architectures from global attention to local attention and three compatibility functions. One of the three compatibility functions was a softmax output of the scalar-products between a target hidden state vector and source hidden state vector. The transformer network \cite{Transformer} uses a similar attention architecture as in \cite{LuongAttention}. Transformer performs a scaled scalar-product between the keys and values and then applies a softmax activation function. Our proposed attention routing computes the scalar product between capsules and a kernel. 

\begin{figure}[t]
\begin{center}
\includegraphics[width=0.95\linewidth]{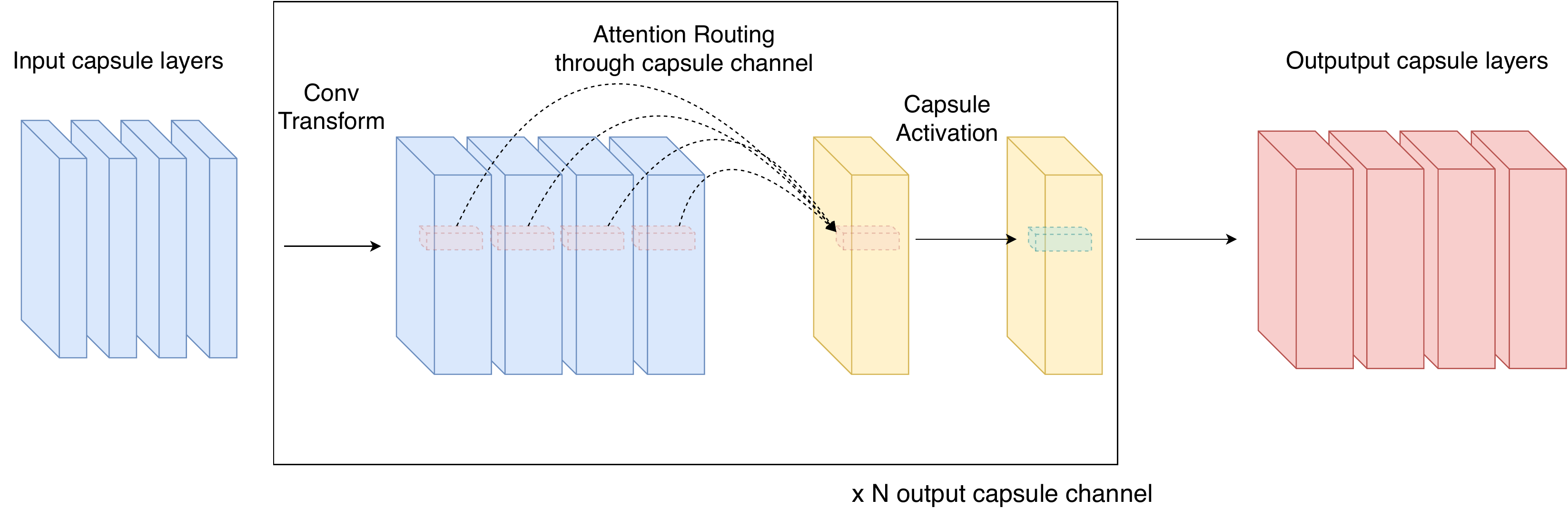}
\end{center}

   \caption{Detailed operation process of conv caps layer. Conv Transform denotes the convolutional transform. The convolutional transform performs a locally connected affine transform on each capsule channel. The attention routing learns the agreement between the convolutional transformed capsules for each spatial location. The capsule activation applies a capsule-wise activation function on each capsule channel.
   }
\label{fig:model}
\end{figure}

\section{Proposed Method}
Our proposed architecture consists of \textit{primary caps layer}, \textit{conv caps layer}, and \textit{fully conv caps layer}.
We denote the $l$th capsule layer as $\mathbf{u}^{l}_{w,h,d,n}$, where $w, h, d$, and $n$ index the spatial width axis, spatial height axis, capsule dimension axis, and capsule channel axis, respectively. We refer to the capsules with the same capsule channel index as a \textit{capsule channel} $\mathbf{u}^{l}_{(:, :, :, n)}$ \footnote{$\mathbf{u}^{l}_{(:, :, :, n_0)} := \{u^{l}_{w,h,d,n}|n=n_0\}$}.

\subsection{Primary Caps Layer}
We denote the primary capsule layer as the $0$th capsule layer. Before entering the primary caps layer, we extract local features $\Tilde{x}$ from the input image $x$ by performing the convolution blocks composed of convolution layer and batch normalization.\cite{BatchNormalization} We consider the local features $\bf{\Tilde{x}}$ as a single capsule layer. In our primary caps layer with N channels of D dimensional output capsules, 3x3 convolution with kernels of filter size D and stride 2 is performed on the input capsules $\bf{\Tilde{x}}$ N times independently. Each output of a convolution layer is a capsule channel.

\begin{equation}
    \mathbf{s}_{(:,:,:,n_0)}^{0} = \textrm{ReLU}\left( \textrm{Conv}_{3\times3} \left( \Tilde{x} \right) \right) 
\end{equation}
Note that this is equivalent to performing a 3x3 convolution of $N\times D$ kernels and then reshaping the features to $(B, W, H, D, N)$ where $B$ denotes the batch size and $(W, H)$ denotes the spatial size of the capsule layer.
Then, the capsule activation is applied to each capsule channel instead of the squash activation function in \cite{capsulenet}.

\subsection{Capsule Activation}
The capsule activation takes an affine transformation on each capsule channel and then applies $\tanh$ activation function. The capsules on the same capsule channel share parameters of the affine transformation. Thus, the capsule activation is equivalent to taking 1x1 convolution with a kernel of filter size D and $\tanh$ activation function on each capsule channel.

\begin{equation}
    \mathbf{u}_{(:,:,:,n_0)}= \tanh\left( \textrm{Conv}_{1\times1} \left( \mathbf{s}_{(:,:,:,n_0)} \right) \right)
\end{equation}
Each element of the output capsules of the capsule activation depends on the corresponding input capsule. Therefore, the capsule activation is a capsule-wise activation function. The $\tanh$ activation function normalizes each element of capsules, thus stabilizes the lengths of the capsules. 

\subsection{Conv Caps Layer}

We denote the input to the $l$th conv caps layer as $\mathbf{u}_{w,h,d,n}^{l-1}$ which is the output of the $(l-1)$th conv caps layer. We first perform a \textit{convolutional transform} on each capsule channel. The convolutional transform is a locally-connected affine transformation sharing parameters within the same capsule channel. In particular, the convolutional transform is a 3x3 convolution of $D^{l}$ kernels without activation function, where $D^{l}$ denotes the capsule dimension of the $l$th conv caps layer. 
\begin{equation}
    \Tilde{\mathbf{s}}_{(:,:,:,m)}^{l, n}  = \textrm{Conv}_{3\times3} \left( \mathbf{u}_{(:,:,:,m)}^{l-1}  \right)
\end{equation}
Each output of the convolutional transform is fed to the attention routing. The output of attention routing is a linear combination of the convolutional transformed capsules with the same spatial location.
\begin{equation}
    \mathbf{s}_{(w,h,:,n)}^{l}  = \sum_{m =1, \cdots, N^{l-1} } c_{(w, h, m)}^{l, n} \cdot \Tilde{\mathbf{s}}^{l, n}_{(w,h,:,m)}
\end{equation}
where the capsules $\mathbf{s}_{(w,h,:, n)}^{l} , \Tilde{\mathbf{s}}^{l}_{(w,h,:,m)}  \in \mathbb{R}^{D^{l}}$. The weights $c_{(w, h, m)}^{l, n} \in \mathbb{R}$ are computed by the attention routing. The log probabilities $b_{(w, h, m)}^{l, n}$ are the scalar-product between a concatenation of capsules $[\Tilde{\mathbf{u}}_{w,h,:,1}^{l}, \Tilde{\mathbf{u}}_{w,h,:,2}^{l}, \cdots, \Tilde{\mathbf{u}}_{w,h,:,N^{l-1}}^{l} ]$ and a parameter vector $w_{n}^{l} \in \mathbb{R}^{D^l \times N^{l-1}}$. This operation can be implemented efficiently by 3D convolution on the convolutional transformed capsule layers with kernels $w_{n}^{l} \in \mathbb{R}^{1 \times 1 \times D^l \times N^{l-1}}$, stride=(1,1,1), and valid padding.

\begin{equation}
	b^{l, n}_{(:, :, :)} =  \textrm{Conv3D}_{1\times1\times D^{l}} \left(  \Tilde{\mathbf{s}}_{(:,:,:,:)}^{l}  \right)
\end{equation}
The weights $c_{(w, h, m)}^{l, n}$ are softmax outputs of the log probabilities $b_{(w, h, m)}^{l, n}$ along the capsule channel axis.
\begin{equation}
	c_{w, h, m}^{l, n} = \frac{ \exp( b_{(w, h, m)}^{l, n} ) }
{ \sum_{1 \leq m \leq N^{l-1}} \exp( b_{(w, h, m)}^{l, n} ) }
\end{equation}
Note that the attention routing adjusts the weight $c_{w, h, m}^{l, n}$ for each spatial location $(w,h)$ corresponding to the convolutional transformed capsules $\{ \Tilde{\mathbf{u}}^{l}_{(w,h,:,m)} \}_{m}$ with the same spatial location.

Finally, the capsule activation is performed on each capsule channel $ \mathbf{s}_{(:,:,:, n)}^{l} $. A set of convolutional transform, attention routing, and capsule activation is performed independently $N^{l}$ times. (\textit{i.e}., each output of the convolutional transform, attention routing, and capsule activation is a capsule channel $  \mathbf{u}_{(:,:,:,n)}^{l}$ )
\begin{equation}
    \mathbf{u}_{(:,:,:,n)}^{l} = \tanh\left( \textrm{Conv}_{1\times1} \left(\mathbf{s}_{(:,:,:,n)}^{l} \right) \right)
\end{equation}


Intuitively, the dynamic routing uses a centroid of the transformed capsules as the reference vector to measure agreement by scalar-product. As the dynamic routing process iterates, the capsule with the higher agreement has a larger weight, and the reference vector evolves in that capsule direction. On the other hand, since the capsule activation in the conv caps layer do not preserve vector orientation, the output capsule $\mathbf{u}_{(:,:,:,n)}^{l}$ cannot approximate the centroid of transformed capsules $\{\Tilde{\mathbf{u}}^{l}_{(w,h,:,n)} \}$. Instead of measuring agreement between the transformed capsules and the output capsule $\mathbf{u}_{(:,:,:,n)}^{l}$, the attention routing parametrizes the routing process. The parameter vector $w_{n}^{l}$ which is the kernel of convolution serves as an approximation of the reference vector to perform routing.

We propose replacing the dynamic routing of \cite{capsulenet} with the convolutional transform and attention routing. Compared to dynamic routing, our proposed operation is faster and more parameter efficient. Since dynamic routing is constructed in a fully connected manner, the transform weight matrices are assigned for each pair of the input capsule and output capsule.
We share the weight matrices across the spatial location and keep the translation equivariance by performing 3x3 convolution on the $l$th layer in the convolutional transform.(Section \ref{Transformation Equivariance}) Besides, the dynamic routing has an iterative routing process to compute the weight $c^{l}_{w, h, n}$. On the other hand, by introducing a trainable parameter vector, our proposed operation is a fast forward-pass.


\subsection{Fully Conv Caps Layer}
The fully conv caps layer is almost the same as the conv caps layer and serves as the output layer of AR CapsNet. The convolutional transform combines capsule features from the all spatial location by applying a kernel of the same spatial size as the input with valid padding. Therefore, the output of the fully conv caps Layer has a shape of $(1, 1, D^{L}, N^{L})$.


\begin{algorithm}[t]
\caption{The process of Attention Routing}

{\bf Input:}  ${\bf u}^{\ell =0} \in  \mathbb{R}^{(w,h,D^{\ell=0},N^{\ell=0})} $ \

\For{$\ell = 1, \cdots, L $}{
    \For{$n = 1, \cdots, N^{\ell}$}{
    \tcc{Convolutional transformation for each capsule channel}
    \For{$m = 1, \cdots, N^{\ell-1}$}{
        $\Tilde{{\bf s }}^{\ell, n}_{(:,:,:,m)} \leftarrow {\tt Conv2D}_{3 \times 3}({\bf u}^{\ell-1}_{(:,:,:,m)})$
        }\ 
    
    \tcc{Attention through capsule channel}
    ${\bf b}^{l, n} \leftarrow {\tt Conv3D}_{1\times1\times D^{\ell}}(\Tilde{{\bf s}}^{\ell})$\ 
    
        \For{$w, h = 1, \cdots, W^{\ell}, H^{\ell}$}{
            ${\bf c}^{n, l}_{w,h,:N^{\ell-1}} \leftarrow {\tt softmax}({\bf b}^{\ell, n}_{w,h,:N^{\ell-1}}) $
            
            ${\bf s}^{\ell}_{w,h,:,n} \leftarrow \sum_{m =1, \cdots, N^{l-1} }{c^{n, l}_{w,h,m} \cdot \Tilde{{\bf s}}^{\ell, n}}_{(w,h,:,m)}$
        }\
    
    \tcc{Capsule activation for each capsule channel}
    ${\bf u }^{\ell}_{(:,:,:,n)}  \leftarrow 
    \tanh({\tt Conv2D}_{1\times1}(\mathbf{s}_{(:,:,:,n)}^{\ell}))$
    }\
    
}

\end{algorithm}

\subsection{Margin Loss and Reconstruction Regularizer} \label{loss}
We adopt the margin loss and reconstruction regularizer in \cite{capsulenet}. Since the output capsules of capsule activation have a length of up to $\sqrt{D^L}$ where $D^L$ denotes the capsule dimension, we use the normalized length to predict the probability of the corresponding class of the dataset.
\begin{equation}
    ||\mathbf{u}^{L}_{n}||_{\mathrm{nor}} = \frac{||\mathbf{u}^{L}_{n}||}{\sqrt{D^L}}
\end{equation}
where $||\mathbf{u}^{L}_{n}||$ denotes the output capsules of the fully conv caps layer and $n$ indexes the capsule channel axis. We applied the Margin loss, $L_n$, for each class $n$ on the $||\mathbf{u}^{L}_{n}||_{\mathrm{nor}}$.
\begin{equation}
\begin{aligned}
    L_n = T_n &\max(0, m^+ - ||\mathbf{u}^{L}_{n}||_{\mathrm{nor}})^2 \\
    &+ \lambda (1-T_n) \max(0, ||\mathbf{u}^{L}_{n}||_{\mathrm{nor}}-m^-)^2
\end{aligned}
\end{equation}
where $T_n = 1$ iff the corresponding class of output capsule is present and $m^+=0.9$ and $m^-=0.1$.

The output capsules $\{\mathbf{u}^{L}_{n}\}_{n=1, \cdots, N}$ are fed to the reconstruction decoder. We used a decoder consisting of 3 fully connected layers as in \cite{capsulenet} except that our decoder has (512, 512, the number of input image pixel) nodes. We refer to the mean of L2 loss between an input image and the decoder output as a reconstruction loss. We add the reconstruction loss that is scaled down by 0.3 to the margin loss as a regularization method.\footnote{CapsuleNet \cite{capsulenet} scaled the reconstruction loss by 0.392. Since we use the mean of L2 loss and CapsuleNet use the sum of L2 loss, 0.392 = 0.0005 $\times$ 784.}

\section{Experiments}

We evaluate our model on the MNIST, affNIST, and CIFAR-10 datasets. For each dataset, we split the training images into a training set (90\%) and a validation set (10\%). We choose the model with the lowest validation error and evaluate the model on the test set. Then, we compare the results with CapsuleNet \cite{capsulenet}.  We use a Keras implementation{\footnote{https://github.com/XifengGuo/CapsNet-Keras}} for CapsuleNet. 

Before training the model on the image dataset, we divide each pixel value by 255 so that it is scaled in the range of 0 to 1. Then, we extract the local features $\Tilde{x}$ from an input image through two convolutional layers with batch normalization(BN) \cite{BatchNormalization} and ReLU activation function. These two convolutional layers use 3x3 kernels with a stride 1. Then, the features go through the AR CapsNet to obtain vector outputs. For each conv caps layer and fully conv caps layer, the dropout layer \cite{Dropout} of keep probability 0.5 is applied to the input capsules before the convolutional transform.

We use the RMSprop optimizer with rho of 0.9 and decay of 1e-4 to minimize the loss defined in Section \ref{loss}. We set the learning rate as 0.001 and batch size as 100

\begin{table*}[t] 
\begin{center}
\begin{tabular}{|l|c|c|c|c|c|}
\hline
Method  & MNIST & MNIST+ & affNIST & C10 & C10+ \\
\hline\hline
CapsuleNet\cite{capsulenet}     & 99.45$^{*}$  & 99.75 (99.52$^{*}$) & 79.0 &  63.1$^{*}$    & 69.6$^{*}$     \\
CapsuleNet+ensemble(7)          & -     & -     & -        & -      & 89.4   \\
Ours                            & 99.46 & 99.46 & 91.6     &  87.19 & 88.61  \\
Ours+ensemble(7)                & -     & -        & -     &  88.94 & 90.11  \\

\hline
\end{tabular}
\end{center}
\caption{
Test accuracy (\%) on the MNIST, affNIST, and CIFAR-10 classification tasks. C10 represents the CIFAR-10 dataset. + denotes training with data augmentation. We adopted translation for MNIST+ and translation, rotation, and horizontal flip for C10+. $^*$ indicates the results from our experiment.
}
\label{table:result_table}
\end{table*}

\subsection{Classification Results on MNIST and affNIST} \label{Results_MNIST_affNIST}

\noindent {\bf Dataset}  \hspace{5pt} 
The MNIST dataset is composed of $28 \times 28$ handwritten digit images. We adopted 0.1 translation as a data augmentation for the MNIST dataset. The affNIST dataset consists of $40\times40$ images, which are obtained by applying various affine transformations such as rotation and expansion to the images from MNIST. For the affNIST classification task, we trained our model with randomly translated MNIST images in horizontal or vertical directions up to shift fraction 0.2 as in \cite{capsulenet}. Any other affine transformations like rotations were not used in the training process. The affNIST dataset has a separate validation set, thus we chose the model with the lowest validation error based on the affNIST validation set. Then, we tested our model with the affNIST test set.

\vspace{5pt} \noindent {\bf Implementation} \hspace{5pt} 
    For the MNIST and affNIST datasets, we used the AR CapsNet which consists of a primary caps layer, one conv caps layer and fully conv caps layer. Before entering the AR CapsNet, an input image goes through two convolutional layers of 64 channels (3x3 Conv - BN - ReLU). The primary caps layer has eight channels of 16-dimensional capsules, the conv caps has eight channels, and the fully conv caps layer has ten channels. Each capsule channels in the conv caps layer and fully conv caps layer has 32 dimensions in the MNIST and 16 dimensions in the affNIST. We decreased the spatial size of the capsule features by applying a 3x3 convolution of stride 2 in the convolutional transform of the conv caps layer. We trained our model for 20 epochs.

\vspace{5pt} \noindent {\bf Accuracy} \hspace{5pt} 
Our model shows a comparable accuracy with the substantial decrease in the number of parameters and training time. Our model with 5.31M parameters achieved 99.45\% accuracy on the MNIST dataset without any data augmentation and 99.46\% accuracy with data augmentation. (Table \ref{table:result_table}) The CapsuleNet with 8.21M parameters achieved 99.45\% accuracy without any data augmentation and 99.52\% with data augmentation. The reported accuracy of CapsuleNet on the MNIST dataset with translation augmentation is 99.75\% \cite{capsulenet}. Also, the training took 37.2 seconds per epoch for our proposed model and 199.5 seconds per epoch for CapsuleNet when we experimented on GTX 1080 GPUs. 

In the affNIST experiments, there are two options to generate training images from the MNIST dataset. The first option is to create a larger dataset by generating a set of all the possible augmented data before training. The second option is to apply translation over the original dataset for each epoch. The reported accuracy of CapsuleNet is 79\% and that of the baseline CNN model is 66\% in \cite{capsulenet}. The experiment is performed on the former option.{\footnote{https://github.com/Sarasra/models/tree/master/research/capsules}} Our proposed model achieved 91.6\% accuracy for the latter option. Under the comparable experiment, our model outperformed the CapsuleNet and the baseline CNN model. Since our model is transformation equivariant (Section \ref{Transformation Equivariance}), our model is robust to affine transformations.

\subsection{Classification Results on CIFAR-10} \label{Results_CIFAR}
\noindent {\bf Dataset}  \hspace{5pt} 
The CIFAR-10 dataset is a $32 \times 32$ colored natural images in 10 classes. We adopted 0.1 translation, rotation up to 20 degrees, and horizontal flip as a data augmentation for CIFAR-10.

\vspace{5pt} \noindent {\bf Implementation} \hspace{5pt} 
For the CIFAR-10 classification task, we added four conv caps layer between a primary caps layer and fully conv caps layer. We decreased the spatial size of the capsule features in the first conv caps layer as in Section \ref{Results_MNIST_affNIST}. Each conv caps layer has eight channels of 32-dimensional capsules and is connected to the next conv caps layer with a residual connection \cite{ResNet}. Note that the residual connection in \cite{ResNet} connects the $l$th layer and $(l+2)$th layer, but our residual connection connects the $l$th conv caps layer and $(l+1)$ conv caps layer. We trained our model for 200 epochs.

\vspace{5pt} \noindent {\bf Accuracy} \hspace{5pt} 
The results in Table \ref{table:result_table} show that our model outperforms CapsuleNet with and without data augmentation. CapsuleNet with 11.74M parameters shows 63.1\% accuracy in C10 and 69.6\% accuracy in C10+. However, our proposed model with 9.6M parameters shows 87.19 \% accuracy in C10 and 88.61\% accuracy in C10+. In \cite{capsulenet}, an ensemble of 7 models achieves 89.4\% accuracy when the models are trained with $24 \times 24$ patches of images and the introduction of a \textit{none-of-the-above} category. However, an ensemble of 7 AR CapsNet models trained with C10+ achieved 90.11\% test accuracy. Note that C10+ only uses rotation, shift, and horizontal flip as data augmentation and not the cropping or the \textit{none-of-the-above} category.

\begin{table}
\begin{center}
\begin{tabular}{|c|c||c|c|c|}
\hline
Conv caps layer  & Caps dim & Params & C10 & C10+\\
\hline\hline
    \multirow{2}{*}{0}  & 16  & 7.3M   & 77.51  & 81.89    \\ 
                        & 32  & 12.6M  & 77.44  & 81.97    \\ \hline
    \multirow{2}{*}{1}  & 16  & 3.5M   & 81.96  & 82.83    \\ 
                        & 32  & 7.7M   & 82.39  & 83.92    \\ \hline
    \multirow{2}{*}{2}  & 16  & 3.7M   & 84.48  & 84.77    \\ 
                        & 32  & 8.4M   & 85.46  & 87.01    \\ \hline
    \multirow{2}{*}{3}  & 16  & 3.8M   & 85.56  & 86.93    \\ 
                        & 32  & 8.9M   & 86.56  & 87.91    \\ \hline
    \multirow{2}{*}{4}  & 16  & 4.0M   & 86.37  & 87.21    \\ 
                        & 32  & 9.6M   & 87.19  & 88.61    \\
\hline 
\end{tabular}
\end{center}
\caption{
Test accuracy (\%) on the MNIST and CIFAR-10 for various hyperparameters. In each experiment, we trained a model for 200 epochs and chose the model with the lowest validation error. For each hyperparameter setting, AR CapsNet shows stable performance without showing severe degradation.
}
\label{table:Hyperparam_variation}
\end{table}

\subsection{Robustness to hyperparameters}

\noindent {\bf Implementation}  \hspace{5pt} 
The AR CapsNet requires a set of hyperparameters, such as the number of conv caps layer and the capsule dimension of each capsule layer. To test the robustness to hyperparameters, we evaluate the AR CapsNet in the CIFAR-10 classification tasks according to the various setting of hyperparameters. 
The evaluated AR CapsNet architecture is the same as the models mentioned in Section \ref{Results_MNIST_affNIST} and \ref{Results_CIFAR}. The primary caps layer has eight channels of 16-dimensional capsules, and the conv caps layer has eight capsule channels. In the setting of hyperparameters, the \textit{Conv caps layer} denotes the number of conv caps layer between the primary caps layer and fully conv caps layer. In every model with at least one conv caps layer, the first conv caps layer decreases the spatial size of the capsule layer by adopting a 3x3 convolution of stride 2 in the convolutional transform. The \textit{Capsule dim} denotes the capsule dimension of the conv caps layer and fully conv caps layer.

\vspace{5pt} \noindent {\bf Robustness} \hspace{5pt} 
All the AR CapsNet models trained with CIFAR-10 dataset show decent performance in Table \ref{table:Hyperparam_variation}. 
Increasing capsule dimension and the number of conv caps layer lead to an improvement in the test accuracy. 
The AR CapsNet model with four conv caps layer shows 86.37\% accuracy with 16-dimensional capsules and 87.19\% accuracy with 32-dimensional capsules. 
The AR CapsNet with four conv caps layer and 32-dimensional capsules shows the best results of 87.19\% in C10 and 88.61\% in C10+.
Also, the AR CapsNet model with no conv caps layer has more parameters than the model with four conv cap layer and shows the worst performance. 
The features in the primary caps layer has a large spatial size. 
Thus, the fully conv caps layer connected to the primary caps layer assigns excessive parameters, and this causes overfitting.


\begin{figure}[t]
\begin{center}
\includegraphics[width=0.95\linewidth]{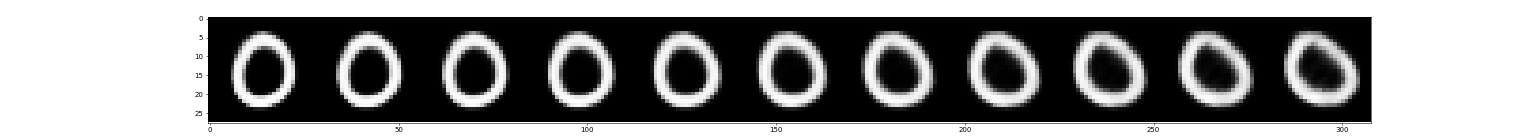}
\includegraphics[width=0.95\linewidth]{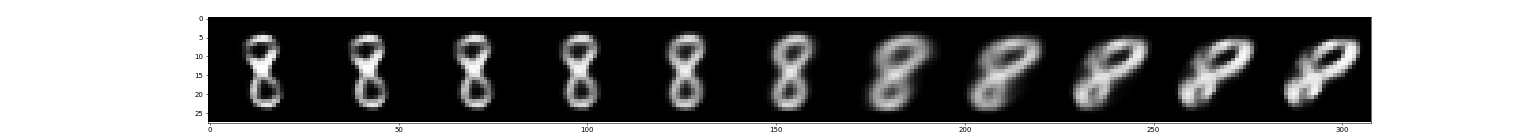}
\includegraphics[width=0.95\linewidth]{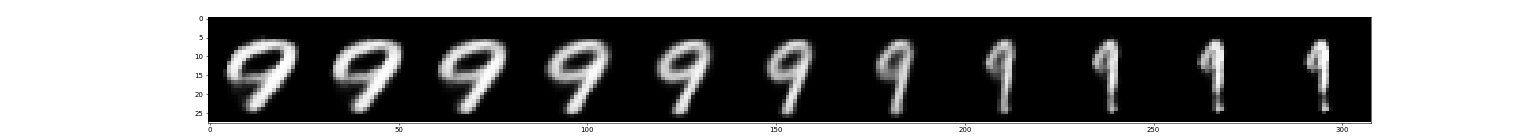}
\end{center}
   \caption{Decoder outputs according to dimension perturbations. We observed the variations of decoder output as we perturbed one dimension of the output capsules by steps of $0.05\sqrt{D^L}$ from $-0.25\sqrt{D^L}$ to $+0.25\sqrt{D^L}$. The perturbation leads to the combination of variations in the decoder output images. (e.g., rotation, thickness, etc.).}
\label{fig:mnist_manipulate}
\end{figure}

\subsection{Transformation Equivariance} \label{Transformation Equivariance}

\noindent {\bf Dimension perturbation}  \hspace{5pt} 
To prove that our proposed model is transformation equivariant, we executed experiments on the MNIST model as in \cite{capsulenet}. We observed the variations of decoder output as we perturbed one scalar element of the output capsules (Figure \ref{fig:mnist_manipulate}). The experiments in the \cite{capsulenet} perturbed one scalar element from -0.25 to 0.25. Since the output capsules of the AR CapsNet have lengths of up to $\sqrt{D^L}$ compared to 1 in \cite{capsulenet}, we perturbed one scalar element from $-0.25\sqrt{D^L}$ to  $0.25 \sqrt{D^L}$ where $D^L$ denotes the capsule dimension of output capsules. Figure \ref{fig:mnist_manipulate} shows that some dimensions of the output capsules represent variations in the way the digit of the corresponding class is instantiated. Some dimensions of the output capsules represent the localized skew in digit 0, the rotation and the size of the higher circle in digit 8, and the rotation, thickness, and skew in digit 9.




\begin{table}
\begin{center}
\begin{tabular}{|l|c|c|c|c|c|c|c|}
\hline
            
Digit&  Rot+&  x+    &  y+    &  Rot-  &  x-    &  y-   \\
\hline\hline
8   &  0.89  &  0.89  &  0.91  &  0.86  &  0.87  &  0.86 \\
5   &  0.89  &  0.78  &  0.73  &  0.84  &  0.88  &  0.86 \\ \hline
\multirow{2}{*}{avg} &  0.88  &  0.86  &  0.84  &  0.83  &  0.83  &  0.84 \\
    &  (0.89)  &  (0.88)  &  (0.80)  &  (0.81)  &  (0.84)  &  (0.84)\\
\hline
\end{tabular}
\end{center}
\caption{
The average of relative ratio $\{r_i\}$ for each combination of digit and transformation. The avg represents the average of all 10,000 test samples for each affine transformation. We report the results of models trained on the MNIST+ dataset in the (). A high relative ratio implies the difference vectors are strongly aligned. For random vectors, the average of relative ratio $\{r_i\}$ is 0.311 and standard deviation is 0.262.
}
\label{table:SVD result table}
\end{table} 

\vspace{5pt} \noindent {\bf Alignment ratio} \hspace{5pt} 
Each scalar element of the output capsules represents a combination of variations such as rotation, thickness, and skew. (Digit 9 in Figure \ref{fig:mnist_manipulate}) Since the length of the output capsules is basis-invariant, the transformation on an input image could be represented in coordinates of any basis. To further test the transformation equivariance of the AR CapsNet, we tested whether the difference in the output capsules caused by applying a transformation on an input image is aligned in one direction. 

Let $\{T_i\}_{i=1, \cdots, N}$ be a set of affine transformations on an input image $x$. We denote the difference between the output capsules $\mathbf{u}_n^{L}(T_i (x))$ and $\mathbf{u}_n^{L}(x)$ as $\mathbf{v}_i(x)$ where n denotes the corresponding class of x.
\begin{equation}
    \mathbf{v}_i(x) = \mathbf{u}_n^{L}(T_i (x)) - \mathbf{u}_n^{L}(x)
\end{equation}
We denote the concatenation of $\mathbf{v}_i(x)$ along the row axis as $\mathbf{V}$. In order to obtain a representative unit vector $\Tilde{v}$ of $\{\mathbf{v}_i(x)\}$, we apply a Singular-Value Decomposition(SVD) on matrix $\mathbf{V}$.
\begin{align}
    \mathbf{c}, \Tilde{v} &= \arg\min_{c_i, \Tilde{v}} \sum_{i} || \mathbf{v}_i(x) - c_{i} \cdot \Tilde{v}  ||_{2}^{2} \\
    &= \arg\min_{c_i, \Tilde{v}} || \mathbf{V} - \mathbf{c} \cdot \Tilde{v}^{T}  ||_{F}^{2}
\end{align}
where F denotes the Frobenius norm, $\mathbf{c} = (c_1, \cdots, c_N)^{T}$, and $c_i \in \mathbb{R}$. The exact solution of this low rank approximation problem is the first right-singular vector $\Tilde{v}$ of $\mathbf{V}$. This experiment is similar to the Principal Component Analysis(PCA) except that we do not subtract the mean for each columns of $\mathbf{V}$. The align vector $\Tilde{v}$ corresponds to the principal vector of PCA. We observed the relative ratio $r_i$ of principal component of $\mathbf{v}_i(x)$ to the vector norms $||\mathbf{v}_i(x)||_2$.
\begin{equation} \label{relative ratio}
    r_i = \frac{|\mathbf{v}_i(x) \cdot \Tilde{v}|}{||\mathbf{v}_i(x)||_2}
\end{equation}


We randomly chose 10,000 images from the test set. For each test image, we generated five images by applying an affine transformation and observed the relative ratio $r_i$. In Table \ref{table:SVD result table}, Rot$(\pm)$ represents $\pm\{5, 10, 15, 20, 25 \}$ degrees rotations and x$(\pm)$ represents a horizontal translation up to $\pm5$ pixels. y$(\pm)$ represents a vertical translation up to $\pm5$ pixels as well. We observed the average of relative ratio $r_i$ for each combination of digit and transformation. Table \ref{table:SVD result table} shows the average of relative ratio $r_i$ for two digits (highest : digit 8, lowest : digit 5) and the average for 10,000 test samples for each transformation. As a reference, we generated five random vectors from the standard multivariate normal distribution. We conducted the same experiment for random vectors for 1,000 times as well. We obtained an average of 0.311 and a standard deviation of 0.262 for random vectors. Even for the worst-case digit 5, every transformation shows a significantly higher relative ratio $r_i$ of 0.73 in y$+$ than the random vectors. This result implies that the difference vectors are strongly aligned in one direction. Therefore, AR CapsNet encodes affine transformations on an input image by some vector components. Also, we report the results of models trained on the MNIST+ dataset in the (). The models trained on the MNIST+ show comparable relative ratio $r_i$ to those trained on the MNIST. This result shows that AR CapsNet encodes affine transformations even without observing transformations during training.

\begin{figure}[t]
\begin{center}
\includegraphics[width=0.8\linewidth]{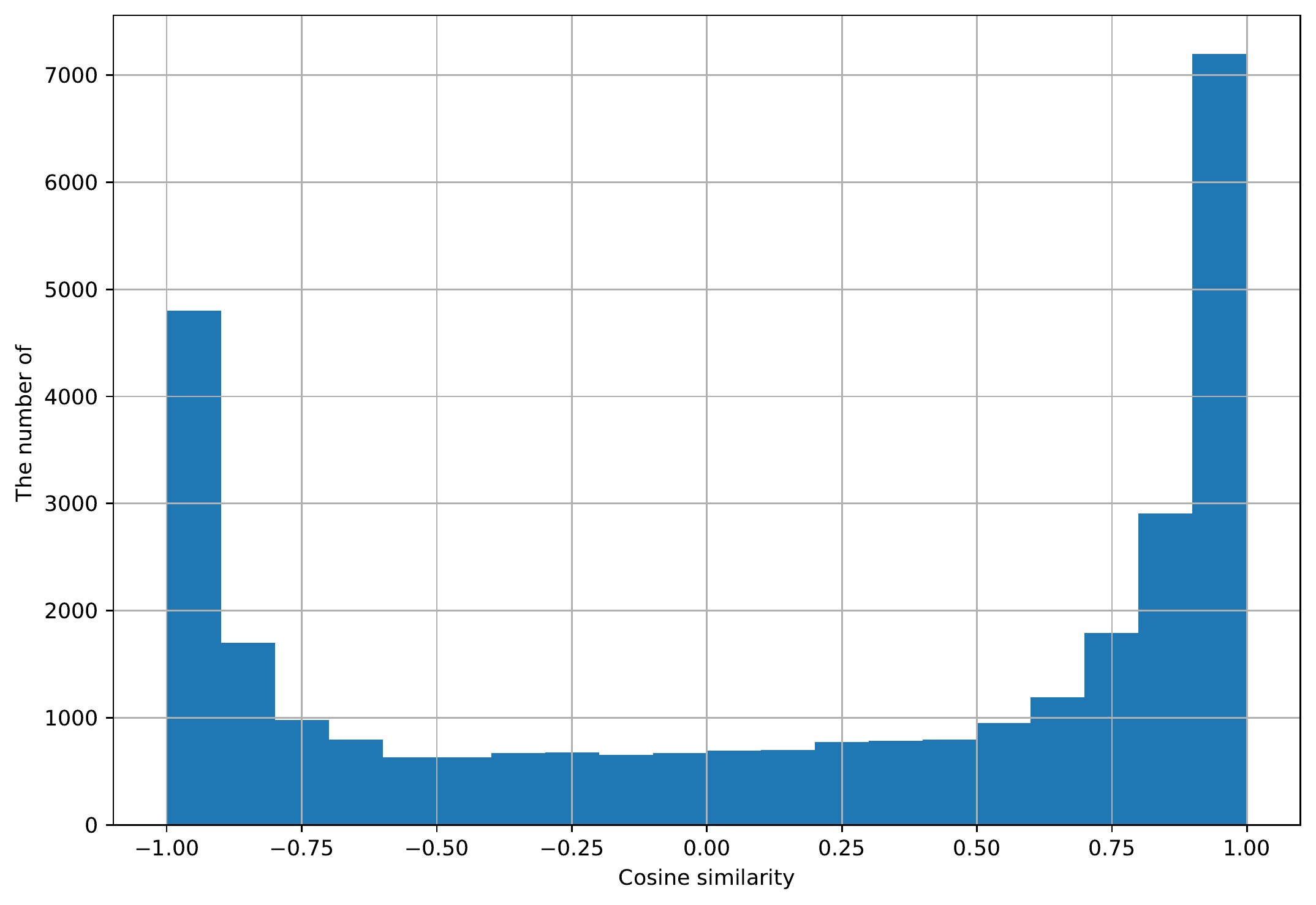}
\end{center}

\caption{Distribution of cosine similarity between unit align vectors $\Tilde{v}$ of positive and negative affine transformations. }
\label{fig:cosine similarity}
\end{figure}

\begin{figure}[t]
\begin{center}
\includegraphics[width=0.85\linewidth]{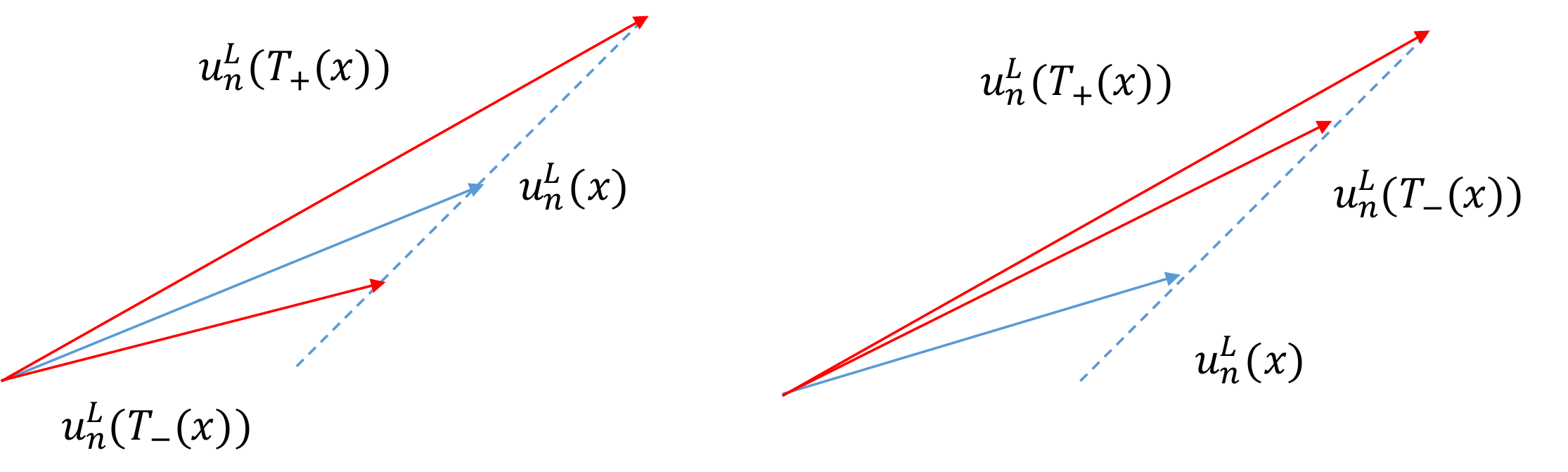}
\end{center}
\caption{Output capsules $\mathbf{u}_n^L$ when the cosine similarity between align vectors of positive and negative transformations is -1 or 1. $T_{+}$ and $T_{-}$ represent the positive and negative transformation. \textbf{Left:} cosine similarity -1. \textbf{Right:} cosine similarity 1. }
\label{fig:output capsule diagram}
\end{figure}

An interesting observation is that AR CapsNet is transformation equivariant but do not distinguish the positive and negative transformations. Figure \ref{fig:cosine similarity} shows the histogram of the cosine similarity between the align vectors of positive and negative transformation.{\footnote{The direction of the first right-singular vector $\Tilde{v}$ is given by $\mathbf{v}_i(x) \cdot \Tilde{v} > 0$ in for each positive and negative transformation. }} We observed two peaks around -1 and 1. The cosine similarity of -1 and 1 imply that positive and negative transformations are encoded in one dimension. However, the cosine similarity of 1 suggests that the difference vectors of positive and negative transformations have the same direction.(Figure \ref{fig:output capsule diagram}) We leave the explanation of this observation to future work.

\section{Conclusion}
In this work, we suggested a new capsule network architecture called Attention Routing CapsuleNet (AR CapsNet). By introducing the attention routing and capsule activation, AR CapsNet obtained a higher accuracy compared to CapsuleNet while using fewer parameters and less training time. 
The attention routing is an effective way to route between capsules because it only compares capsules of the same spatial location. In addition, the attention routing does not require an iterative routing process as the dynamic routing does because it directly learns the weights between capsules. The capsule activation is based on the assumption that the capsule-scale activation can extract transformation equivariant features even if it is not orientation-preserving. This assumption distinguish the capsule activation from the squash activation function and its variant. 

While using the building blocks of CNNs, AR CapsNet is transformation equivariant. We showed that capsules have transformation information by manipulating the output capsules and then observing the decoder output images. Also, we observed the difference vectors between the output capsules of an original image and an affine transformed image. By showing that the difference vectors are strongly aligned in one direction, we proved that AR CapsNet encodes transformation information in some dimensions. There are natural variations of AR CapsNet such as introducing a feature compression by 1x1 convolution to the capsule activation and a transformer network \cite{Transformer} to the attention routing. We plan to study these variations in the future.

\newpage

{\small
\bibliographystyle{ieee}
\bibliography{egbib}
}

\end{document}